%% file: conference_101719.tex
\documentclass[conference]{IEEEtran}
\IEEEoverridecommandlockouts
\usepackage{cite}
\usepackage{amsmath,amssymb,amsfonts}
\usepackage{algorithmic}
\usepackage{graphicx}
\usepackage{textcomp}
\usepackage{float}
\usepackage{multirow}
\usepackage{xcolor}
\def\BibTeX{{\rm B\kern-.05em{\sc i\kern-.025em b}\kern-.08em
    T\kern-.1667em\lower.7ex\hbox{E}\kern-.125emX}}
\begin{document}

\title{HyT-NAS: Hybrid Transformers Neural Architecture Search for Edge Devices\\
}

\author{\IEEEauthorblockN{Lotfi Abdelkrim Mecharbat\IEEEauthorrefmark{1},
Hadjer Benmeziane\IEEEauthorrefmark{2}, Hamza Ouarnoughi\IEEEauthorrefmark{2} and Smail Niar\IEEEauthorrefmark{2} }
\IEEEauthorrefmark{1}Ecole Nationale Supérieure d'Informatique, Algiers, Algeria \\
\IEEEauthorrefmark{2}Université Polytechnique Hauts-de-France, Valenciennes, France \\
Email: 
\IEEEauthorrefmark{1}hl\_mecharbat@esi.dz \\
\IEEEauthorrefmark{2}\{firstname.lastname\}@uphf.fr
}

\maketitle

\begin{abstract}
Vision Transformers have enabled recent attention-based Deep Learning (DL) architectures to achieve remarkable results in Computer Vision (CV) tasks.
However, due to the extensive computational resources required, these architectures are rarely implemented on resource-constrained platforms.
Current research investigates hybrid handcrafted convolution-based and attention-based models for CV tasks such as image classification and object detection.
In this paper, we propose HyT-NAS, an efficient Hardware-aware Neural Architecture Search (HW-NAS) including hybrid architectures targeting vision tasks on tiny devices.
HyT-NAS improves state-of-the-art HW-NAS by enriching the search space and enhancing the search strategy as well as the performance predictors.
Our experiments show that HyT-NAS achieves a similar hypervolume with less than ~5x training evaluations.
Our resulting architecture outperforms MLPerf MobileNetV1 by 6.3\% accuracy improvement with 3.5x less number of parameters on Visual Wake Words.
\end{abstract}

\begin{IEEEkeywords}
component, formatting, style, styling, insert
\end{IEEEkeywords}
\section{Introduction}
\label{sec:introduction}
\input{sections/introduction}

\section{Background \& Related Works}
\label{sec:related_works}
\input{sections/related_works}
\vspace{-0.5cm}
\section{Proposed Approach}
\label{sec:approach}
\input{sections/approach}
\vspace{-0.25cm}
\subsection{Hybrid Attention-based Search Space}
\label{sec:search_space}
\input{sections/search_space}

\subsection{Search Algorithm}
\label{sec:search_strategy}
\input{sections/search_strategy}

\subsection{Accuracy \& Latency Predictors}
\label{sec:predictors}
\input{sections/predictors}

\section{Evaluation}
\label{sec:experiments}
\input{sections/experiments}

\section{Conclusion}
This paper introduces HyT-NAS, a Hardware-aware Neural Architecture Search (HW-NAS) for hybrid attention and convolution based models, targeting edge and tiny devices.
We fulfilled the initial promise of bringing hybrid models to the tiny realm with less than 300k parameters and state-of-the-art accuracy for Visual Wake Words and Person Detection.

\newpage
\bibliographystyle{plain}
\bibliography{biblio}

\end{document}

%% file: sections/introduction.tex
Recent advances have proven the efficacy of the attention operation, in different domains such as Natural Language Processing (NLP) or Computer Vision (CV). 
For the latter, Vision Transformers (ViT)~\cite{DBLP:conf/iclr/DosovitskiyB0WZ21} outperform state-of-the-art in image  classification~\cite{DBLP:conf/nips/DaiLLT21} and object detection~\cite{DBLP:journals/corr/abs-2203-03605}. 

In parallel, there is a growing need of vision tasks deployment at the edge for privacy and efficiency purposes.
The remarkable success of ViT architectures is often at the expense of a huge number of parameters and a time-consuming inference. 
This makes ViT unsuitable for edge and tiny devices. 

\begin{figure}
    \centering
    \includegraphics[trim=0 0 1cm 2cm, clip, 
    width=0.5\textwidth]{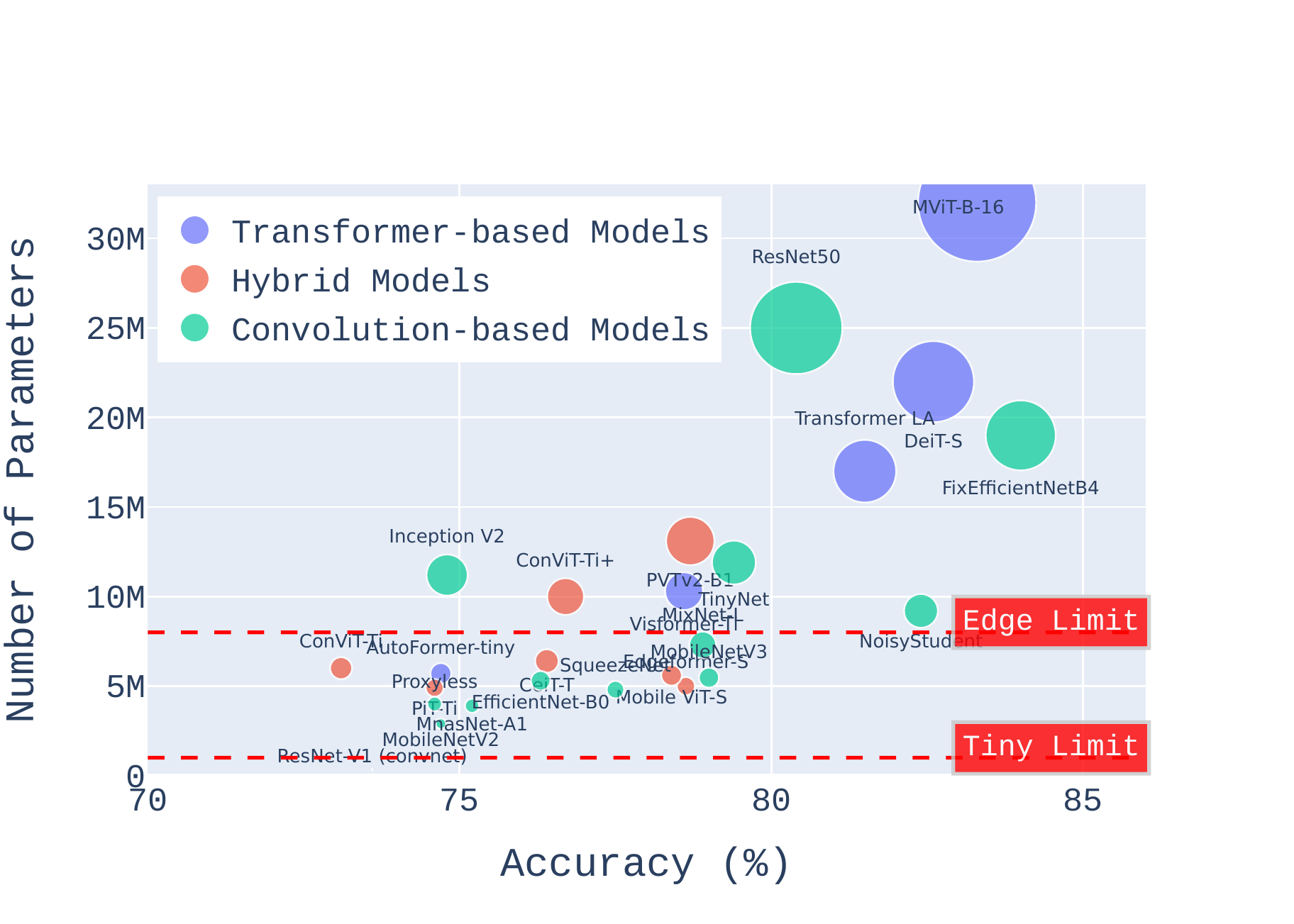}
    \caption{Number of parameters and accuracy on Imagenet of state-of-the-art DL architectures. The limits are obtained from MCUNet~\cite{DBLP:conf/nips/LinCLCG020}}
    
    \label{fig:1}
\end{figure}

Current research investigates the hybrid convolution and attention architectures to meet the edge requirements. 
Attention operations are mainly used to enhance the task-specific performance, e.g., accuracy for image classification or the Average Precision (AP) for object detection. 
However, even hybrid networks~\cite{convit, pit} are far from an optimal edge deployment.  
Figure~\ref{fig:1} shows the difference in terms of number of parameters and accuracy of state-of-the-art architectures on image classification. 
We notice that under the edge limit, almost all architectures are convolution-only with some hybrid architecture. 
In addition, near the tiny limit which allows up to 1M parameters, all the architectures are convolution-only. 

\begin{figure*}[!ht]
    \centering    \includegraphics[width=0.85\textwidth]{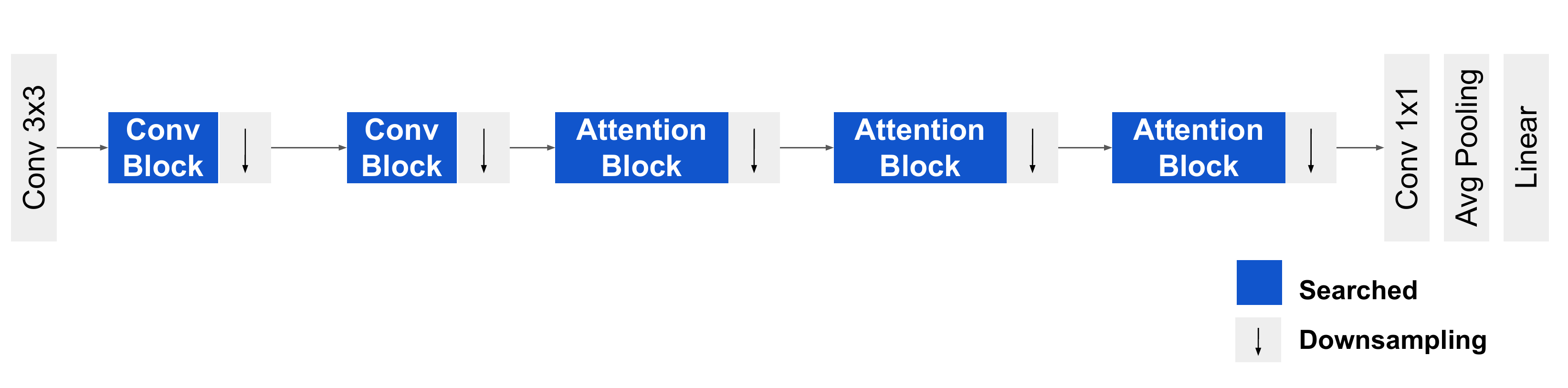}
    \caption{Hybrid attention and convolution based macro-architecture}
    \label{fig:2}
\end{figure*}
Hardware-aware Neural Architecture Search (HW-NAS)~\cite{survey} automates the identification of efficient neural network (NN) architecture given a task-specific evaluation criterion and hardware-specific constraints, such as latency and energy consumption.
The problem is then cast to a multi-objective optimization search where the obtained result is a Pareto front. 
The latter represents a set of NN architectures that give the best trade-off between the different objectives. 

In this paper, we propose HyT-NAS, an optimized HW-NAS for hybrid attention and convolution-based architectures. 
Our contributions can be summarized as follows:
\begin{enumerate}
    \item A new hybrid search space that includes convolution and attention blocks targeting tiny and edge devices. 
    \item An adapted search strategy based on Multi-Objective Bayesian Optimization using a customized performance evaluation predictors. 
    \item HyT-NAS library enables to use the search methodology for a given task and hardware platform defined by the user. Our code and library are available~\footnote{https://anonymous.4open.science/r/HyT-NAS-Search-Algorithm-A864/README.md}.  
\end{enumerate}
Using HyT-NAS, we bring down the size of hybrid architectures to the tiny architectures while achieving interesting performances. 
Our final architectures have been tested on Visual Wake Words and Person Detection. The best obtained architecture outperforms MLPerf MobileNetV1 by +6.3\% more accuracy with 3.5x less number of parameters on Visual Wake Words, while on Person Detection, we achieve the same mAP as MobileNetV3 with a 7x smaller model.

The rest of the paper is structured as follows: First, in section \ref{sec:related_works} we give an overview of Hardware-aware Neural Architecture Search and Hybrid Attention and Convolution Architectures. 
Section \ref{sec:approach} details our approach HyT-NAS with its components. 
Finally, we provide extensive comparison between HyT-NAS and state-of-the-art search methodologies and apply our final models in Visual Wake Words and Object Detection. The results are presented in section~\ref{sec:experiments}.

%% file: sections/related_works.tex

\subsection{Hardware-aware Neural Architecture Search}
HW-NAS~\cite{survey} is cast as a multi-objective optimization problem where the search space represents a set of possible architectures. Given that, increasing the accuracy tends to increase the model's size and thus its latency and energy consumption, all the objectives conflict. The results in a Pareto front that contains the most interesting solutions in term of trade-off between the different objectives.

Evaluating the accuracy and hardware performances for each sampled architecture hinders HW-NAS. Several methods~\cite{DBLP:conf/nips/WhiteZRLH21} estimate the accuracy. A surrogate model is built to predict the architecture's ranking using the learning to rank theory~\cite{DBLP:conf/icml/WistubaP20, DBLP:conf/arabwic/BenmezianeOMN21} based on its characteristics. A common practice to assess the search strategies is to compare the number of evaluations required by each algorithm. 


\subsection{Bayesian Optimization}
Bayesian Optimization (BO) is a sample-efficient methodology to optimize expensive black-box functions such as the accuracy that requires hours of training. Intuitively BO tries to answer the question: \textit{Based on what we know so far, which point should we evaluate next in the search process?}. It leverages a probabilistic surrogate model, generally a Gaussian process~\cite{10.5555/3221315.3221581}. BO build an acquisition function to enable exploring uncertain regions that might contain good solutions. Acquisition functions are heuristics that provide the  desirable it is to evaluate a point. 
In multi-objective settings, the acquisition function~\cite{10.5555/3221315.3221581} tries to improve the hypervolume of the Pareto front approximation in each step. 
\vspace{-0.25cm}
\subsection{Hybrid Attention and Convolution Architectures}

Vision transformers~\cite{DBLP:conf/iclr/DosovitskiyB0WZ21} have made a remarkable success in Computer vision outperforming Convnets in several tasks such as image classification~\cite{DBLP:conf/nips/DaiLLT21} and object detection~\cite{DBLP:journals/corr/abs-2203-03605}.
Many papers~\cite{DBLP:conf/nips/RaghuUKZD21},\cite{tuli2021convolutional} compared the representations extracted from a ConvNet and Transformers. A general result is that both operations extract different features, giving rise to hybrid attention and convolution networks.

ConViT~\cite{convit} modifies the attention operation. It incorporates soft convolutional inductive biases using a gated positional self-attention. PiT~\cite{pit} merges Vit and ConvNets by inserting a depth-wise convolution-based pooling layer in the original ViT definition. 

%% file: sections/approach.tex
This paper describes a dedicated search methodology for tiny and edge hybrid attention and convolution architectures. This methodology is composed of two fundamental components: the \textit{Hybrid Attention-Convolution Search Space} and the \textit{HyT-NAS Search strategy}.


%% file: sections/search_space.tex
Our search space is inspired by the MobileVit~\cite{mobilevit} architecture. Figure~\ref{fig:2} depicts the macro-architecture used to construct all architectures in our search space. The architecture consists of a series of interleaved convolution and transformer blocks. Each block has its own set of hyperparameters described in Table~\ref{tab:1}. The total size of our search space is approximately $1.2 * 10^{9}$.

Each convolution block is a MobileNetV2~\cite{mobilenetv2} block. We sweep over: the number of convolutions, their expansion ratio, and their output channel size. These blocks' inner output channel size is then computed with the input channel size multiplied by the expansion ratio. Each attention block is a Vision transformer encoder block. Particularly, we sweep the number of heads and MLP layers size contained in each attention block. 

\begin{table}[h!]
\begin{tabular}{|c|p{3cm}|c|}
\hline
\textbf{Block}                            & \textbf{Hyperparameter}                                       & \textbf{Values}              \\ \hline
\multirow{3}{*}{\textbf{Convolution Block}}       & \textbf{Number of blocks}                                     & \textbf{{[}1, 2, 3, 4{]}}   \\ \cline{2-3} 
                                           & \textbf{Expand ratio}                    & \textbf{{[}1x, 2x, 4x{]}}      \\ \cline{2-3} 
                                           & \textbf{Out channel size}                                     & \textbf{{[}8, 16, 24, 32{]}} \\ \hline
\multirow{5}{*}{\textbf{Attention Block}} & 
     \textbf{Expand ratio}                    & \textbf{{[}1x, 2x, 4x{]}}      \\ \cline{2-3} 
                                           & \textbf{Channel size}                                         & \textbf{{[}1x, 1.5x, 2x{]}}      \\ \cline{2-3} 
                                           & \textbf{Number of  heads}                                     & \textbf{{[}1, 2, 4{]}}      \\ \cline{2-3} 
                                           & \textbf{Feed forward ratio} & \textbf{{[}1x, 1.5x,  2x{]}}   \\ \hline
\end{tabular}
\caption{Search space hyperparameters}
\label{tab:1}
\vspace{-0.5cm}
\end{table}

%% file: sections/search_strategy.tex

Our search strategy is an optimized multi-objective bayesian optimization (MOBO). 

\begin{figure}[t]
    \centering
    \includegraphics[width=.5\textwidth]{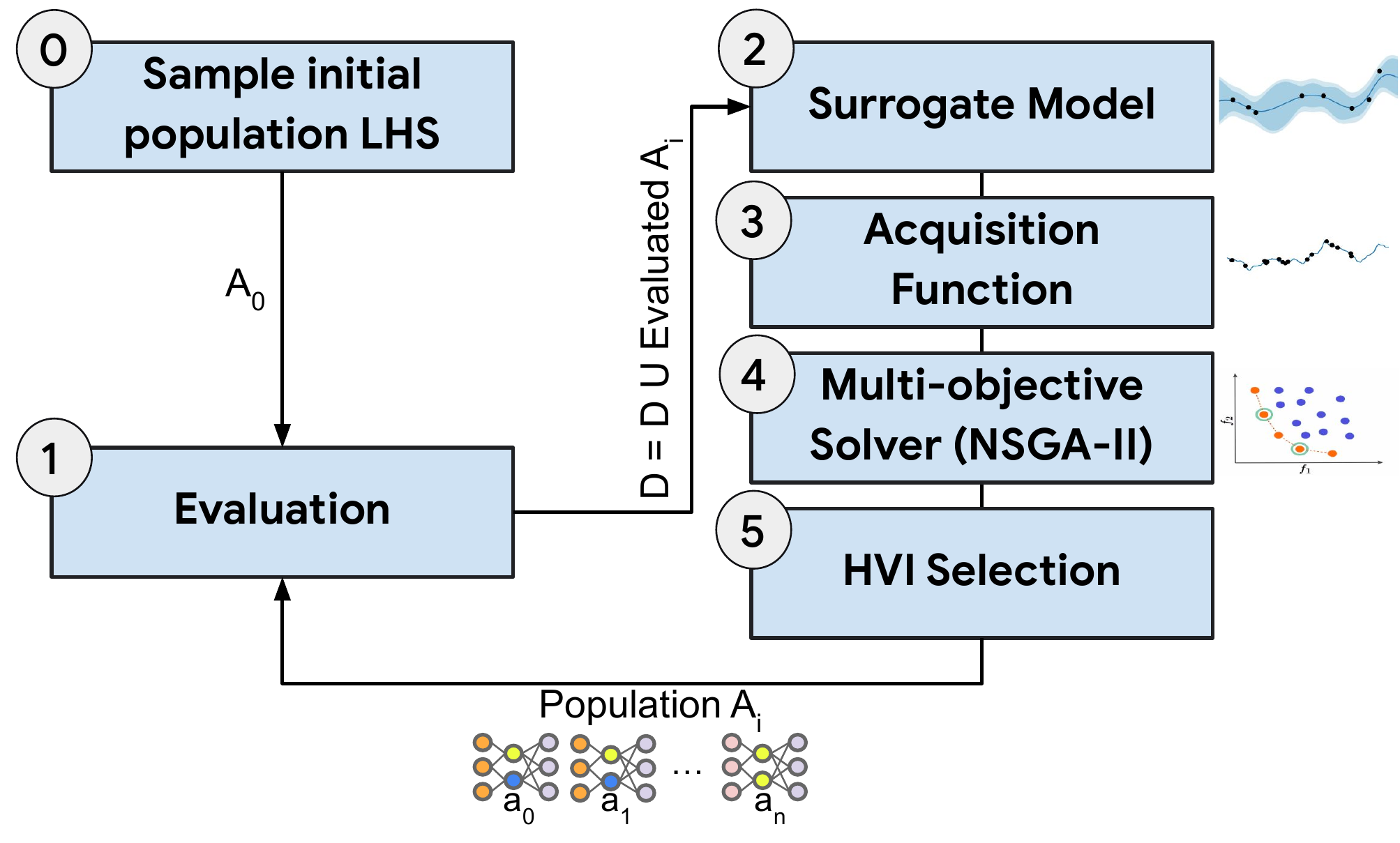}
    \caption{HyT-NAS Search Strategy }
    \label{fig:search_algo}
    \vspace{-0.5cm}
\end{figure}

Figure~\ref{fig:search_algo} shows the pipeline of our search algorithm. Steps 1 to 5 are repeated for a user-defined number of iterations:
\begin{itemize}
    \item First, an initial population of architectures $A_{0}=\{ a_1,...,a_n \} $ is sampled via latent hypercube sampling~\cite{DBLP:journals/technometrics/McKayBC00}. The elements of the population  will then be evaluated to construct a dataset $D={(a_1,y_1),...,(a_n,y_n)}$ where  $y_i$ is a tuple containing the accuracy and latency of the architecture $a_i$. The dataset  grows incrementally throughout the search.
    \item In each iteration, the surrogate model is trained using the dataset containing all previously evaluated points. The predictions of the model are then used to approximate the objectives via an acquisition function.
    \item Finally, the surrogate problem defined as minimizing the acquisition function is solved using NSGA-II. The optimal points found by the solver will go through a selection process based on Hypervolume Improvement (HVI). This selection will result in the construction of the new population $A_i$ that will be evaluated and added to the dataset $D$.
\end{itemize}

\textit{Surrogate model:} In the standard MOBO, a Gaussian process (GP) is used to model each objective independently. GPs are known for their low performance on high dimensional data (d$>$10) and their poor scalability with respect to the number of evaluations. 
 Thus, we replace them with a set of rank predictors of different depths and widths. 
Rank predictors are models that rank the values of the target instead of approximating them. These models are trained at each iteration on all previously evaluated architectures $D$. The mean and standard deviation of the networks' predictions are used to approximate the rank of the objectives via an acquisition function.

\textit{Acquisition function:} The most common acquisition functions used in literature are: expected improvement (EI), probability of improvement (PI), and upper confidence bound (UCB). The latter is used in our search strategy. The UCB algorithm enables us to balance exploration and exploitation. It shifts from concentrating on exploration, choosing the least preferred actions, to focusing on exploitation, which is adequate to our surrogate model's learning.

\textit{Selection strategy:} Our selection strategy involves two steps: First, points obtained from the solver are sorted using the non-dominated sorting algorithm. Then, the points belonging to the two first dominance levels are passed to the second selection phase based on HVI.

%% file: sections/predictors.tex

To reduce the search time, we build ML predictors to predict the architecture's accuracy and latency within hundreds of milliseconds, which takes the search time a few seconds. We uniformly sample architectures from our search space to construct the training dataset. 



The latency dataset is more extensive for two reasons: (1) the latency increase is subtle from one architecture to the other, provided we do not increase the number of blocks. In other terms, if we only increase the expansion ratio or the channel size, the increase in the latency is small but not negligible. The predictor thus requires more training points to catch these differences. (2) It takes more time to train the architectures than to run them on the target hardware. 

We empirically tested multiple ML predictors: XGBoost, LGBoost, MLP (2layers) and MLP (3layers). Each predictor was trained to predict the accuracy and latency using the mean squared error. We compared the kendal tau ranking correlation of each predictor. XGBoost gives 95\% and 93\% for latency and accuracy estimation respectively. 

%% file: sections/experiments.tex
\subsection{Expirements setup and Implementation Details}
\textit{Search Algorithm settings:} We initialize each ranker randomly from a set of predefined hyperparmaters which are: number of estimators=\{100, 400, 800, 1000\}, max depth for the estimators =\{3, 6, 12\} and learning rate =\{0.01, 0.1, 0.5\}. We used NSGA-II with a population size of 100 and a number of generations of 20.

\textit{Evaluation strategy settings:} we have used predictors for both objectives : accuracy and latency. These predictors were built using a dataset of 300 models trained partially for 50 epochs on 8 Nvidia A100 GPUs accessed through Grid5000\cite{grid5000} for accuracy, and a dataset of 1700 models executed each using an image size of 224 on a Raspberry Pi Model 3 B For latency. 

\begin{table*}[h!]
\centering
\begin{tabular}{|p{3cm}|l|c|l|c|}
\hline
\textbf{Model}           & \textbf{Accuracy (\%)} & \textbf{Latency (s)} & \textbf{Nb Param (M)} & \textbf{\begin{tabular}[c]{@{}c@{}}Hardware-aware \\ Search\end{tabular}} \\ \hline
MobileNetV1              & 83.7                   & 2.61                    & 0.67                  & No                                                                        \\ \hline
MobileNetV2 (x0.35)      & 86.34                  & 4.23                   & 1.7                   & No                                                                        \\ \hline
ProxylessNAS             & 86.55                  & 2.51                    & 4.0                   & Yes                                                                       \\ \hline
MobileVit-XS             & 82.14                  & 2.08                   & 2.3                   & No                                                                        \\ \hline
MobileVit-S              & 84.64                  & 2.49                & 5.6                   & No                                                                        \\ \hline
\textbf{HyT-NAS-BL (ours)} & \textbf{85.48}         & \textbf{0.47}       & \textbf{0.015}        & \textbf{Yes}                                                              \\ \hline
\textbf{HyT-NAS-BA (ours)} & \textbf{92.25}         & \textbf{2.04}      & \textbf{2.7}          & \textbf{Yes}                                                              \\ \hline
\textbf{HyT-NAS-O (ours)}  & \textbf{90.02}         & \textbf{0.63}       & \textbf{0.187}        & \textbf{Yes}                                                              \\ \hline
\end{tabular}
\caption{Comparison between state-of-the-art architectures and HyT-NAS optimal architectures on Visual Wake Words. BL stands for "Best Latency". BA stands for "Best Accuracy". "O" stands for optimal trade-off.}
\label{tab:result}
\vspace{-0.75cm}
\end{table*}
\subsection{End-to-end Results}

We compare our search strategy to three state-of-the-art and well-known optimization algorithms: Random Search (RS), NSGA II, and Multi-objective Bayesian Optimization (MOBO). For a fair comparison, we use our latency and accuracy predictors for each method, i.e., the evaluation is done by calling the two predictors. 

\begin{figure}
    \centering
    \includegraphics[width=0.45\textwidth]{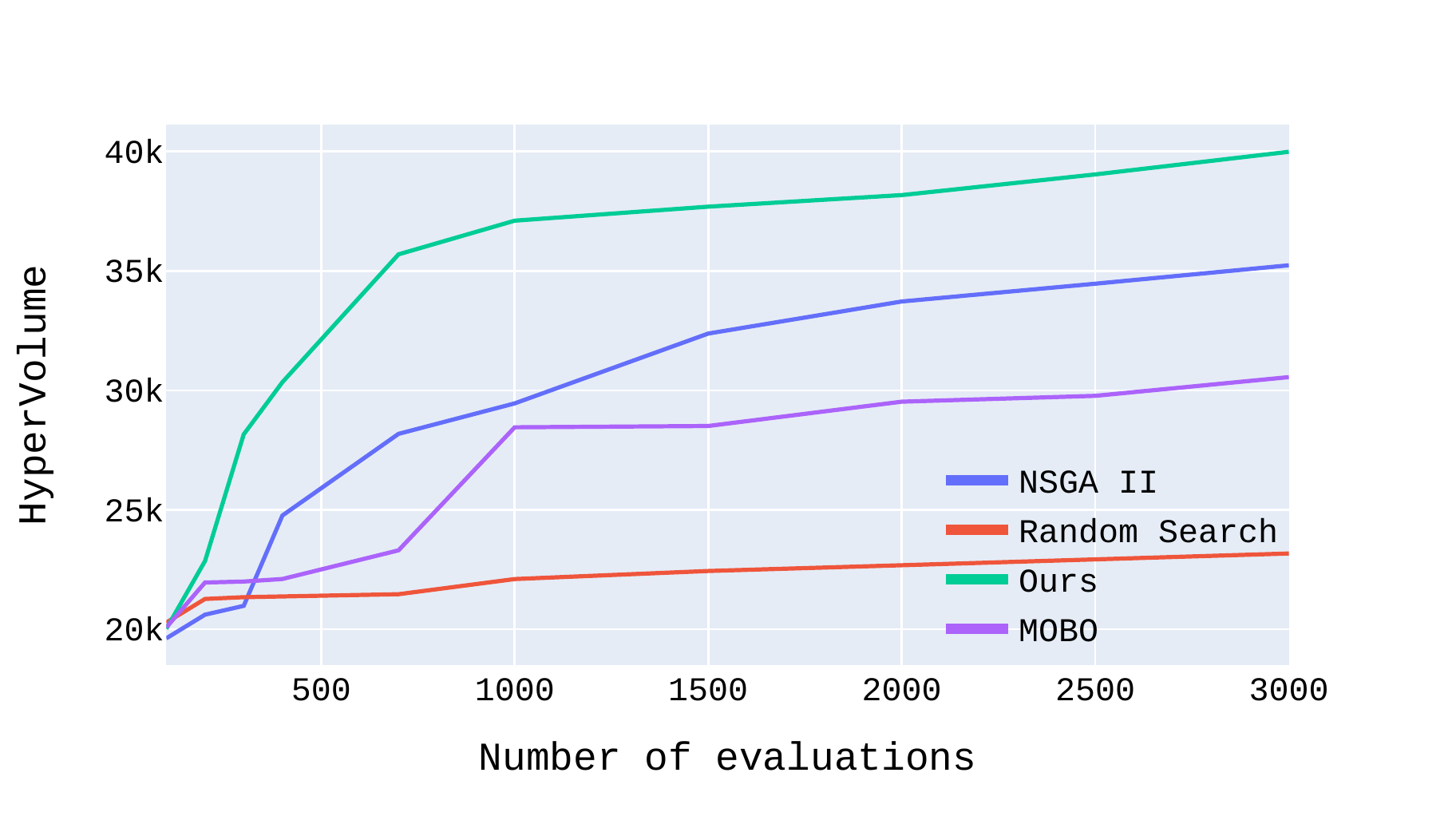}
    \caption{Evolution of the hypervolume with respect to the number of evaluations for state-of-the-art search strategies}
    \label{fig:search_evolution}
    \vspace{-0.5cm}
\end{figure}

\begin{figure}
    \centering
    \includegraphics[width=0.45\textwidth]{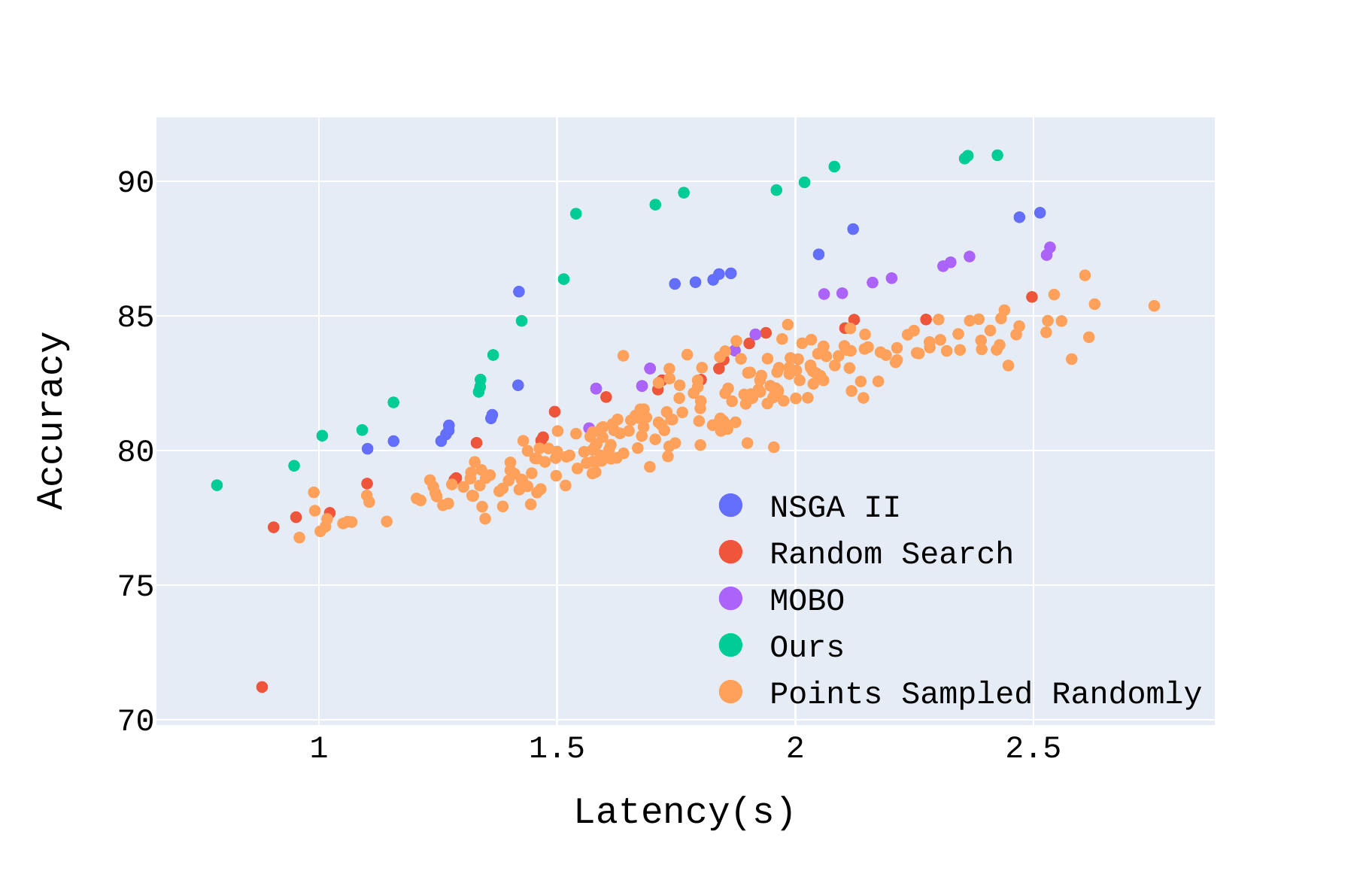}
    \caption{Pareto front approximations using different search strategies}
    \label{fig:pareto_fronts}
    \vspace{-0.75cm}
\end{figure}

Figure~\ref{fig:search_evolution} shows the evolution of the hypervolume with respect to the number of evaluations. Compared to these methods, our strategy rapidly achieves a higher hypervolume with less than 500 evaluations. We can notice the importance and efficiency of optimizing the surrogate model in MOBO compared to GP. Figure~\ref{fig:pareto_fronts} shows the Pareto front approximations. Our method, in green, get the closest to the optimal corner, i.e., where the latency is minimal and the accuracy is maximal. 

Overall our search strategy takes few minutes to achieve a better Pareto front approximation. Our work targets tiny DL architectures. For this reason, we first target Visual wake words task. This task consists on classifying whether or not an image contains a person.

Table~\ref{tab:result} compares state-of-the-art architectures to HyT-NAS optimal architectures. From the final Pareto front approximation, we select three architectures: \textit{HyT-NAS-BL} corresponding to the architecture with minimum latency, \textit{HyT-NAS-BA} corresponding to the architecture with maximum accuracy, and \textit{HyT-NAS-O} an in-between architecture. HyT-NAS-BL outperforms MobileVit variants while significantly decreasing the latency and number of parameters. This architecture only contains 15k parameters, which makes even its training faster. HyT-NAS-BA outperforms all state-of-the-art architectures in terms of accuracy and latency while having a decent number of parameters under the edge limit. Our optimal architecture, HyT-NAS-O, decreases the number of parameters to 187k, making it suitable for tiny devices. HyT-NAS-O outperforms state-of-the-art architectures in terms of accuracy. 




\subsection{Use Case: Object Detection }\label{sec:od}
Object detection is an important task at the edge. It is used in autonomous driving, robots, and medical assistants~\cite{DBLP:conf/icra/Munoz-MartinezVGFSGTJ00}. Generally, object detection models are composed of three components: Backbone, Neck and Head.
Typically, the backbone is the most time-consuming component among almost all state-of-the-art models. This why it's the focus of our optimization method.



We apply our optimal model as a backbone and use SSD-Lite~\cite{DBLP:conf/eccv/LiuAESRFB16} as head. We compare our optimal architecture with state-of-the-art models. For a fair comparison, we use the same training methodology explained in MobileVit~\cite{mobilevit} to fairly compare the models. Table\ref{tab:odresults} illustrates that our model is 3.8x smaller with 2.8\% off from the medium MobileVit-XS. Compared to MobileNetV3, we achieve the same mAP with a 7x smaller model.  

\begin{table}[H]
    \centering
    \begin{tabular}{p{3cm}|c|c}
    \hline
         Backbone & Nb param & mAP \\ \hline
         MobilenetV3 & 4.9M & 22.0 \\
         MobileViT-XXS & 1.9M& 21.4 \\ 
         MobileViT-XS  & 2.7M & 24.8 \\
         HyT-NAS-O (ours) & 0.7M & 22.0\\ \hline
    \end{tabular}
    \caption{Object Detection with SSD-Lite}
    \label{tab:odresults}
\end{table}
